%% file: neruips_2023.tex
\documentclass{article}

\usepackage[numbers]{natbib}
\usepackage[preprint]{neurips_2023}

\usepackage[dvipsnames]{xcolor}         
\definecolor{linkColor}{rgb}{0.18,0.39,0.62}
\usepackage[utf8]{inputenc} 
\usepackage[T1]{fontenc}    
\usepackage[colorlinks=true,linkcolor=linkColor,citecolor=linkColor,filecolor=linkColor,urlcolor=linkColor]{hyperref}       
\usepackage{url}            
\usepackage{booktabs}       
\usepackage{amsfonts}       
\usepackage{nicefrac}       
\usepackage{microtype}      
\usepackage{xcolor}         
\usepackage{natbib}
\usepackage[utf8]{inputenc} 
\usepackage[T1]{fontenc}    
\usepackage{booktabs}       
\usepackage{amsfonts}       
\usepackage{nicefrac}       
\usepackage{microtype}      
\usepackage{xcolor}         
\usepackage{pifont}
\usepackage{threeparttable}
\usepackage{multirow}
\usepackage{amsfonts}
\usepackage{bbm}
\usepackage{bm}
\usepackage{bbold}
\usepackage{color}
\usepackage{graphicx}
\usepackage{amsmath}
\usepackage{subfigure}
\usepackage{enumitem}
\usepackage[noend]{algorithm2e}
\usepackage{wrapfig}
\input{Definitions}
\newcommand\our{\makebox{\textsc{LongMem}}}
\allowdisplaybreaks[4]

\title{Augmenting Language Models with \\ Long-Term Memory}

\author{Weizhi Wang$^\dagger$,~~Li Dong$^\ddagger$,~~Hao Cheng$^\ddagger$,~~Xiaodong Liu$^\ddagger$,\\
\textbf{Xifeng Yan$^\dagger$,~~Jianfeng Gao$^\ddagger$,~~Furu Wei$^\ddagger$}  \\
$^\dagger$University of California, Santa Barbara \ \ 
$^\ddagger$Microsoft Research \\
\texttt{weizhiwang@ucsb.edu,~\{lidong1, haocheng\}@microsoft.com}\\
}

\begin{document}

\maketitle

\input{0_abstract}
\input{1_introduction}

\input{2_method}
\input{3_experiment}
\input{4_related_work}

\input{5_conclusion}



{\small
\bibliography{sidenet}
}

\newpage
\appendix
\input{supplementary}


\end{document}

%% file: 0_abstract.tex
\begin{abstract}
Existing large language models (LLMs) can only afford fix-sized inputs due to the input length limit, preventing them from utilizing rich long-context information from past inputs.
To address this, we propose a framework, Language Models Augmented with \textbf{Long}-Term \textbf{Mem}ory (\our{}), which enables LLMs to memorize long history. We design a novel decoupled network architecture with the original backbone LLM frozen as a memory encoder and an adaptive residual side-network as a memory retriever and reader. Such a decoupled memory design can easily cache and update long-term past contexts for memory retrieval without suffering from memory staleness.
Enhanced with memory-augmented adaptation training, \our{} can thus memorize long past context and use long-term memory for language modeling.
The proposed memory retrieval module can handle unlimited-length context in its memory bank to benefit various downstream tasks. Typically, \our{} can enlarge the long-form memory to 65k tokens and thus cache many-shot extra demonstration examples as long-form memory for in-context learning. 
Experiments show that our method outperforms strong long-context models on ChapterBreak, a challenging long-context modeling benchmark, and achieves remarkable improvements on memory-augmented in-context learning over LLMs.
The results demonstrate that the proposed method is effective in helping language models to memorize and utilize long-form contents. Our code is open-sourced at \url{https://aka.ms/LongMem}.

\end{abstract}

%% file: 1_introduction.tex
\section{Introduction}
\label{sec:intro}
Large language models (LLMs) have revolutionized natural language processing with great successes in advancing the state-of-the-art on various understanding and generation tasks \citep{bert,gpt2,roberta,Yang2019XLNetGA,gpt3,t5}.
Most LLMs benefit from self-supervised training over large corpora via harvesting knowledge from fix-sized local context, showing emergent abilities, \eg zero-shot prompting~\citep{gpt2}, in-context learning~\citep{gpt3}, and Chain-of-Thought (CoT) reasoning~\citep{cot}.
Nevertheless, the input length limit of existing LLMs prevents them from generalizing to real-world scenarios where the capability of processing long-form information beyond a fix-sized session is critical, \eg long horizontal planning.

\begin{figure*}[!t] 
\centering 
\includegraphics[width=0.7\textwidth]{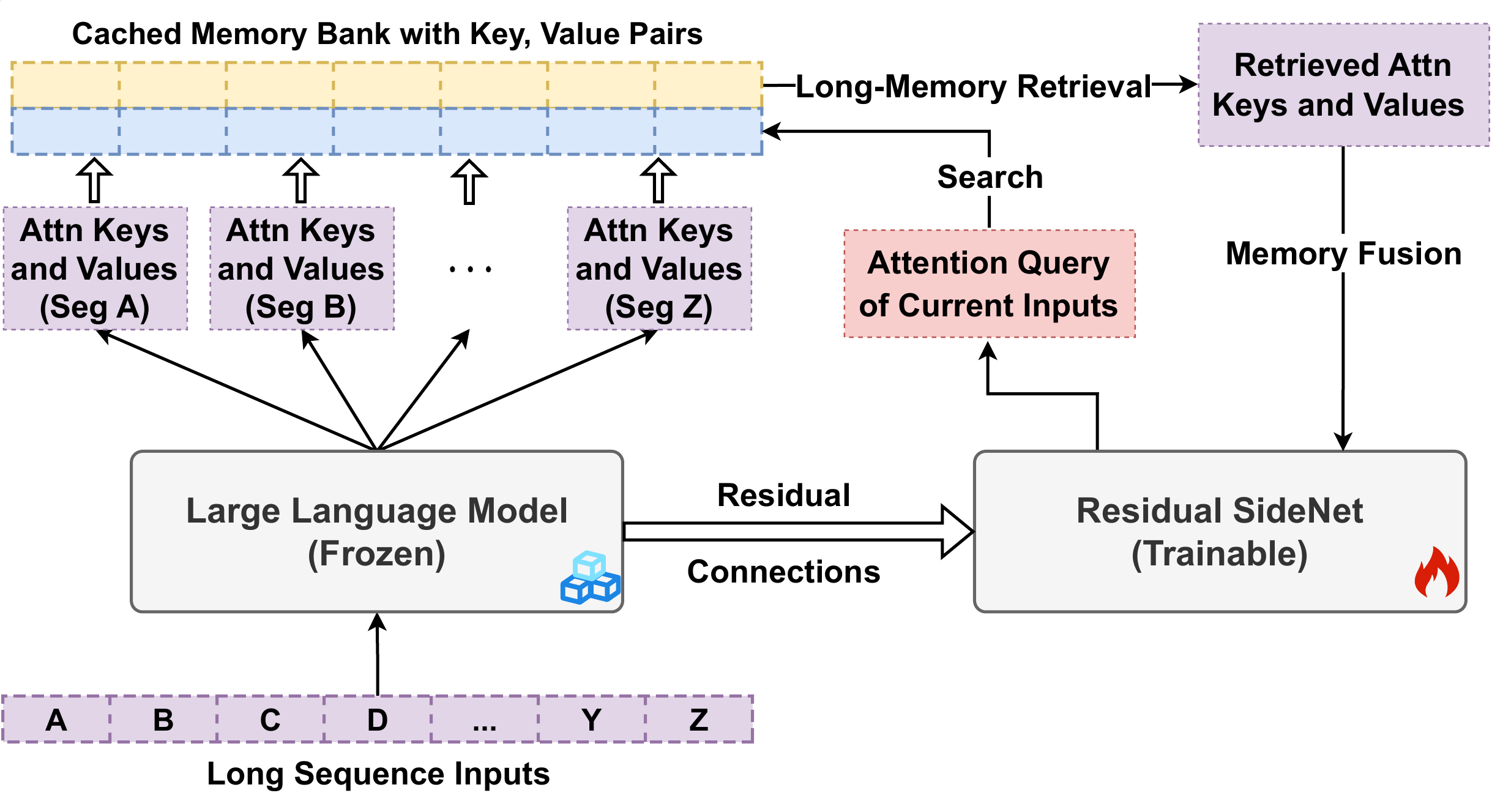} 
\caption{Overview of the memory caching and retrieval flow of \our{}. The long text sequence is split into fix-length segments, then each segment is forwarded through large language models and the attention key and value vectors of $m$-th layer are cached into the long-term memory bank. For future inputs, via attention query-key based retrieval, the top-$k$ attention key-value pairs of long-term memory are retrieved and fused into language modeling.
}
\vspace{-10pt}
\label{fig:mem}
\end{figure*}

To address the length limit issue, the most straightforward method is to simply scale up the input context length.
For instance, GPT-3~\citep{gpt3} increases the input length from 1k of GPT-2~\citep{gpt2} to 2k tokens for capturing better long-range dependencies.
However, this approach typically incurs computation-intensive training from scratch and the \textit{in-context} dense attention is still heavily constrained by the quadratic computation complexity of Transformer self-attention~\citep{Vaswani2017AttentionIA}.
Another recent line of work \cite{beltagy2020longformer,bigbird} instead focuses on developing in-context sparse attention to avoid the quadratic cost of self-attention, which still largely requires training from scratch.
In contrast, the prominent work, Memorizing Transformer (MemTRM)~\citep{Wu2022MemorizingT}, approximates in-context sparse attention via dense attention over both in-context tokens and memorized tokens retrieved from a non-differentiable memory for Transformers.
Thus, MemTRM scales up the resulting language model to handle up to 65k tokens and achieves substantial perplexity gains in modeling full-length books or long papers.
However, MemTRM faces the \textit{memory staleness} challenge during training due to its coupled memory design, which uses a single model for encoding memory and fusing memory for language modeling.
In other words, as the model parameters are updated, cached older representations in memory may have distributional shifts from those from the latest model, thereby limiting the effectiveness of the memory augmentation.

In this paper, we propose a framework for Language Models Augmented with \textbf{Long}-Term \textbf{Mem}ory (\our{}), which enables language models to cache long-form previous context or knowledge into the non-differentiable memory bank, and further take advantage of them via a decoupled memory module to address the memory staleness problem.
To achieve decoupled memory, we design a novel residual side-network (SideNet).
Paired attention keys and values of the previous context are extracted using a frozen backbone LLM into the memory bank.
In the memory-augmented layer of the SideNet, the generated attention query of the current input is used to retrieve cached (keys, values) of previous contexts from the memory, and the corresponding memory augmentations are then fused into learned hidden states via a joint-attention mechanism.
Furthermore, newly designed cross-network residual connections between the SideNet and the frozen backbone LLM enable better knowledge transfer from the pretrained backbone LLM.
By continually training the residual SideNet to retrieve and fuse memory-augmented long-context, the pre-trained LLM can be adapted to leverage long-contextual memory for improved modeling. The detailed memory cache, retrieval and fusion process is illustrated in Figure~\ref{fig:mem}.

Our decoupled memory design leads to two main benefits.
First, our proposed architecture decouples the process of encoding previous inputs into memory and the process of memory retrieval and fusion by decoupled frozen backbone LLM and SideNet.
In this way, the backbone LLM only works as the long-context knowledge encoder, while the residual SideNet works as the memory retriever and reader, which effectively resolves the issue of memory staleness.
Second, directly adapting the entire LLM with memory augmentations is computationally inefficient, and also suffers from catastrophic forgetting.
As the backbone LLM is frozen during the efficient memory-augmented adaptation stage, \our{} can not only tap into the pretrained knowledge but also avoid catastrophic forgetting. 

\our{} is capable of taking various types of long-form text and knowledge into the memory bank based on downstream tasks.
Here, we consider two representative cases, language modeling with full-length book contexts, and memory-augmented in-context learning with thousands of task-relevant demonstration examples. 
Specifically, we evaluate the effectiveness of the proposed \our{} on various long-text language modeling, and memory-augmented in-context learning for language understanding. 
Experimental results demonstrate that our model consistently outperforms the strong baselines in terms of long-text modeling and in-context learning abilities.
Our method substantially improves LLM's long-context language modeling capabilities by -1.38$\sim$-1.62 perplexity over different length splits of Gutenberg-2022 corpus.
Remarkably, our model achieves the state-of-the-art performance of 40.5\% identification accuracy on ChapterBreak, a challenging long-context modeling benchmark, significantly surpassing existing strong x-former baselines. 
Lastly, with 2k demonstration examples in memory, \our{} 
 shows pronounced in-context learning improvements on popular NLU tasks, compared with MemTRM and non-memory-augmented baselines.

%% file: 2_method.tex

\section{Methods}
\label{sec:methods}

To enable LLMs to harvest relevant information from the past long context in memory, we propose to augment the frozen backbone LLM with a decoupled memory module. 
To fuse the memory context information, we design a novel lightweight residual SideNet, which can be continually trained in an efficient way.
In the following, we first discuss the problem formulation of language modeling with memory augmentations.
Then, we formally introduce our efficient residual SideNet for adapting the frozen pretrained LLM to jointly attend over local input context and retrieved memory context.
Lastly, we provide our designed processes of how past memory is encoded, stored, recalled and fused for language modeling. 

\begin{figure*}[t] 
\centering 
\includegraphics[width=\textwidth]{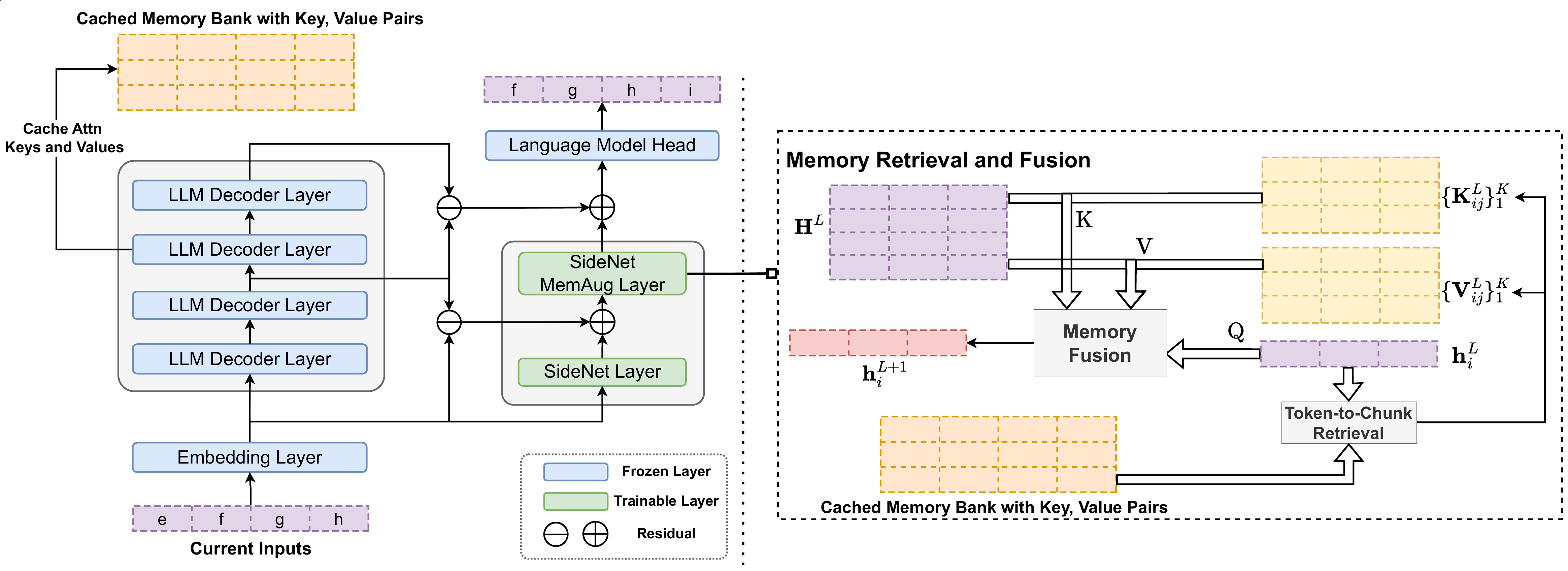} 
\caption{Overview of \our{} architecture. ``MemAug'' represents Memory-Augmented Layer.
}
\label{fig:model}
\end{figure*}

\subsection{Language Models Augmented with Long-Term Memory}
\label{ssec:mem_llm}
Here, we focus on the high-level problem setup and defer more component details to later sections. Given its wide adoption for pretrained LLMs, our \our{} model is built on the Transformer architecture \cite{Vaswani2017AttentionIA}.
For \our{}, there are three key components: the frozen backbone LLM, SideNet, and Cache Memory Bank.
As most existing pretrained LLMs can only take a fix-sized input, only the input segment of a long sequence (\eg a book) that can fit in the length limit is denoted as the current input as done for most existing autoregressive language models.
Those previous segments that can not fit are denoted as previous inputs, which are used for memory augmentations.
To tap into the learned knowledge of the pretrained LLM, both previous and current inputs are encoded using the frozen backbone LLM but different representations are extracted. 
For previous inputs, the key-value pairs from the Transformer self-attention at $m$-th layer are stored in Cache Memory Bank, whereas
the hidden states from each LLM decoder layer for the current inputs are retained and transferred to SideNet.
For each current input token, top relevant key-value vector pairs are retrieved as memory augmentations for language modeling. 
The SideNet module can be viewed as an efficient adaption model that is trained to fuse the current input context and relevant cached previous contexts in the decoupled memory.

Formally, for a fix-sized input text sequence $\{\mathbf{x}_i\}_{i=1}^{|x|}$ (the current input), \our{} first performs a forward pass using the backbone LLM (marked in Blue in Figure~\ref{fig:model}) \textbf{without any gradient calculation}.
The embedding layer of the backbone LLM first encodes the input $\{\mathbf{x}_i\}_{i=1}^{|x|}$ into embedding space and outputs the initial hidden states, $\mathbf{H}^0_\text{LLM}\in\RR^{|x|\times E}$, where $E$ is the hidden dimension.
Then each successive Transformer decoder layer of the frozen backbone LLM computes the new hidden states using the hidden states from the previous layer, $\mathbf{H}_{\text{LLM}}^{l^\prime}=f_{\theta_{\textnormal{LLM}}^{l^{\prime}}} (\mathbf{H}_{\text{LLM}}^{l^{\prime}-1}), \forall l^{\prime}\in[1,{L^\prime}]$ and ${L^\prime}$ is the total \# layers for the backbone LLM.
During the forward pass with the backbone LLM for all previous inputs, the key-value pairs used for self-attention at the $m$-th Transformer decoder layer are stored in \textbf{Cached Memory Bank} (marked in Orange in Upper-Left corner of Figure\ref{fig:model}), which are later recalled as memory augmentations for future inputs. 

\textbf{Cached Memory Bank} is a cached head-wise vector queue $\mathcal{Z}_{k},\mathcal{Z}_{v}\in \mathbb{R}^{H \times M\times d}$, which maintains attention key-value pairs of latest $M$ previous inputs $\widetilde{\mathbf{K}},\widetilde{\mathbf{V}}\in \mathbb{R}^{H\times |x|\times d}$, where $H, d$ denotes the number of attention heads and per-head dimension respectively. After memory retrieval and fusion (\S\ref{ssec:mem_rf}), the memory bank removes the key-value pairs of the oldest sequences and appends the current sequences to the cached vector bank.
Thus such an update mechanism ensures the language modeling causality at the sequences level and enables the memory bank to always keep records of the nearest previous context for the current inputs.

After the forward pass with the backbone LLM, the SideNet module then takes all current input hidden states from the backbone LLM $\{\mathbf{H}_\text{LLM}^{l^\prime}\}_{{l^\prime}=1}^{L^\prime}$ and the past key-value pairs in Cached Memory Bank for computing memory-augmented representations.
Specifically, our SideNet of \our{} consists of $(L-1)$ normal Transformer decoder layers and one special memory-augmented decoder layer.
For efficient purposes, we mainly consider the case where \#layers $L$ of the SideNet is smaller than that of the backbone LLM, \ie $L < {L^\prime}$.
Our SideNet encodes $\mathbf{H}^0$ into memory-augmented contextual representation via $(L-1)$ normal Transformer decoder layers and a special \textbf{memory-augmented layer}.

The \textbf{memory-augmented layer} is an extension of the vanilla Transformer decoder layer that takes a memory-augmented input, including both top relevant key-value pairs in memory and the hidden states from the current input.
Here, the cached key-value pairs are recalled using a token-based memory retrieval module (\S\ref{ssec:mem_rf}).
For each current input token, the memory retrieval module $s_{rt}(:)$ retrieves top-$K$ relevant key-value pairs in the memory bank $\{\widetilde{\kvec}_{ij},\widetilde{\vvec}_{ij} \}_{j=1}^K=s_{rt}(\mathbf{x}_i)$.
Then SideNet computes the output using the memory-augmented input, $\mathbf{H}^{m_s}_{\text{Side}}=f_{\theta_{\textnormal{Mem}}}(\mathbf{H}_{\text{Side}}^{m_s-1}, \{\{\widetilde{\mathbf{k}}_{ij},\widetilde{\mathbf{v}}_{ij}\} _{j=1}^K \}_{i=1}^{|x|})$, where $m_s$ is the layer index where we inject the memory-augmentation layer.

Finally, the token probability is computed using the last SideNet hidden states 
$P(\mathbf{x}_i|\mathbf{x}_1,\cdots,\mathbf{x}_{i-1})=\textnormal{softmax}(W\mathbf{H}^L)$, where $W$ is the frozen output embedding weight shared by both the backbone LLM and SideNet. We perform a memory-augmented adaptation training for \our{} to utilize the decoupled memory. Following the \textit{generative unsupervised pre-training}~\cite{gpt1}, the training objective of \our{} is the standard left-to-right language modeling objective, which maximizes the likelihood of the next token based on the left context: $\max \sum_{x \in \mathcal{D}} \sum_{i=1}^{|\mathbf{x}|} \log P(\mathbf{x}_i|\mathbf{x}_1,\cdots,\mathbf{x}_{i-1}),$
where $x$ is a randomly sampled sentence from the pre-training text corpus $\mathcal{D}$.

\subsection{Residual SideNet}
\label{ssec:res_sidenet}
\textbf{SideNet Architecture and Initialization.} Here, we again implement SideNet based on Transformer~\cite{Vaswani2017AttentionIA}.
Here, the number of decoder layers $L$ in SideNet is equal to the number of layers $L^{\prime}$ in the backbone LLM divided by a reduction factor (a layer reduction factor of $2$ throughout this work $L^\prime=2L$).
The weights of each decoder layer in SideNet are initialized from the corresponding pre-trained decoder layer of the backbone LLM with the same depth: $\Theta_{\textnormal{Side}}^{{l^\prime}\over 2}=\Theta_{\textnormal{LLM}}^{{l^\prime}}$.
As illustrated in Figure~\ref{fig:model}, the SideNet takes the output of backbone LLM's embedding layer and reuses the language modeling head layer of backbone LLM, which is also frozen during the continual adaption stage.
During the memory-augmented adaptation stage, all other parameters of SideNet are updated accordingly based on the training signal. In this way, the lightweight SideNet achieves fast convergence with knowledge transferred from pre-trained parameters.

\textbf{Cross-Network Residual Connections.} 
To tap into knowledge from the pretrained backbone LLM, we resort to proposed cross-network residual connections for fusing representations from the backbone LLM into SideNet. Specifically, we add the difference between output hidden states at $2l$-th and $(2l-2)$-th layers of the backbone LLM as the residual connections to the output hidden states at $l$-th layer of SideNet.
Then, the input to the next $(l+1)$-th layer of SideNet is the sum of the original hidden state forwarded through the previous layer $f_{\Theta_{\textnormal{Side}}^l}(\mathbf{H}_\text{Side}^{l-1})$ and the cross-network residual connection of the hidden state difference from the backbone LLM 
\begin{align}
\mathbf{H}^{l}_{\textnormal{Side}}=f_{\Theta^l_{\textnormal{Side}}}(\mathbf{H}_{\textnormal{Side}}^{l-1})+(\mathbf{H}_{\textnormal{LLM}}^{2l}-\mathbf{H}_{\textnormal{LLM}}^{2l-2}),\forall l\in [1,L],
\label{eq:residual}
\end{align} 
where $\mathbf{H}^0$ is the output of embedding layer. It is worth noting that the residual connections after the self-attention and feed-forward network of a decoder layer~\cite{Vaswani2017AttentionIA} will be performed as normal in $f_{\Theta^l_{\textnormal{Side}}}(\mathbf{H}_{\textnormal{Side}}^{l-1})$ and parallel to the proposed cross-network residual connections. 


\subsection{Memory Retrieval and Fusion}
\label{ssec:mem_rf}
The long-term memory capability of \our{} is achieved via a memory-augmentation module for retrieval and fusion.

\textbf{Token-to-Chunk Memory Retrieval.}
Instead of performing token-to-token retrieval, we focus on token-to-chunk retrieval for acceleration and integrity. A text-chunk refers to an n-gram structure of chunk-size $csz$ number of contiguous tokens. The memory bank stores cached key-value pairs at the level of token chunks. We divide the memory bank into $M/csz$ attention key-value paired chunks and use the mean-pooled vector on the chunk-size dimension to get the key vector for retrieval.
Then we retrieve the top-$(K/csz)$ attention key-value chunks w.r.t the dot product between the attention query of the current input token and the mean-pooled attention key of a candidate chunk. Finally, we squeeze the chunk-size dimension for retrieved key-value paired chunks and flatten them into $K$ key-value pairs at token-level $\{\widetilde{\mathbf{K}}_{j},\widetilde{\mathbf{V}}_{j} \}_{j=1}^K$.
Adopting token-to-chunk retrieval reduces the size of the retrieval index and accelerates the process. Meanwhile, the retrieval accuracy can be further improved, which is also observed in \cite{n-gram-nnmt} and \cite{RETRO}. The hyperparameter chunk-size $csz$ controls the granularity of retrieved contexts, which can be empirically adjusted based on downstream tasks.
For instance, in-context learning requires more fine-grained label tokens from demonstration examples cached in memory, where a smaller $csz$ is helpful.

\textbf{Memory Fusion.}
The memory fusion is performed within a special memory-augmented layer. As the conventional Transformer decoder layer uses the multi-head self-attention~\cite{Vaswani2017AttentionIA}, we follow \cite{Wu2022MemorizingT} to extend it to a joint-attention mechanism and propose a long-term memory fusion process to enable each token to attend on both local contexts and retrieved memory contexts. With the head-wise hidden state output from previous layer $\mathbf{H}^{l-1}\in\mathbb{R}^{|x|\times d}$ and the corresponding retrieved attention key-value pairs are $\{\widetilde{\mathbf{K}}_{i},\widetilde{\mathbf{V}}_{i}\}_{i=1}^{|x|}\in \mathbb{R}^{|x|\times\mathrm{K}\times\mathrm{d}} $, the output hidden state for the $l$-th memory-augmented layer $\mathbf{H}^l$ is computed as:
\begin{align}
&\mathbf{A} = \textnormal{softmax}(\frac{\mathbf{Q}\mathbf{K}^T}{\sqrt{d}})\mathbf{V}, \ \ \mathbf{M} = \textnormal{Concat}\{\textnormal{softmax}(\frac{\mathbf{Q}_i \widetilde{\mathbf{K}}_{i}^T}{\sqrt{d}})\widetilde{\mathbf{V}}_{i}\}_{i=1}^{|x|}, \\
&\mathbf{H}^l = \textnormal{sigmoid}(g)\cdot \mathbf{A} + (1-\textnormal{sigmoid}(g))\cdot \mathbf{M},
\end{align}
where $\mathbf{Q},\mathbf{K}, \mathbf{V},\mathbf{A},\mathbf{M}\in \mathbb{R}^{|x|\times \mathrm{d}}$, $\mathrm{K}$ is the number of retrieved attention key-value pairs in cached memory for each token, and $g$ is a trainable head-wise gating vector. The hidden state output from previous layer $\mathbf{H}^{(l-1)}$ is linearly projected into attention queries, keys, and values $\mathbf{Q},\mathbf{K},\mathbf{V}$ separately via three matrices $W^Q, W^K, W^V\in \mathbb{R}^{\mathrm{d}\times \mathrm{d}}$. It is worth noting that the retrieved attention key-value pairs in cached memory are distinct to each token.

%% file: 3_experiment.tex
\section{Experiments}
\label{sec:exp}

We evaluate our proposed \our{} model on different tasks based on the demanded in-memory long-contexts:
a) long-text language modeling and language understanding when loading the past long-context into cached memory;
b) infinite-length in-context learning when loading large number of demonstration examples into cached memory.

\subsection{Training Setup}
\label{sec:setup}
\begin{figure*}[ht] 
\centering 
\includegraphics[width=\textwidth]{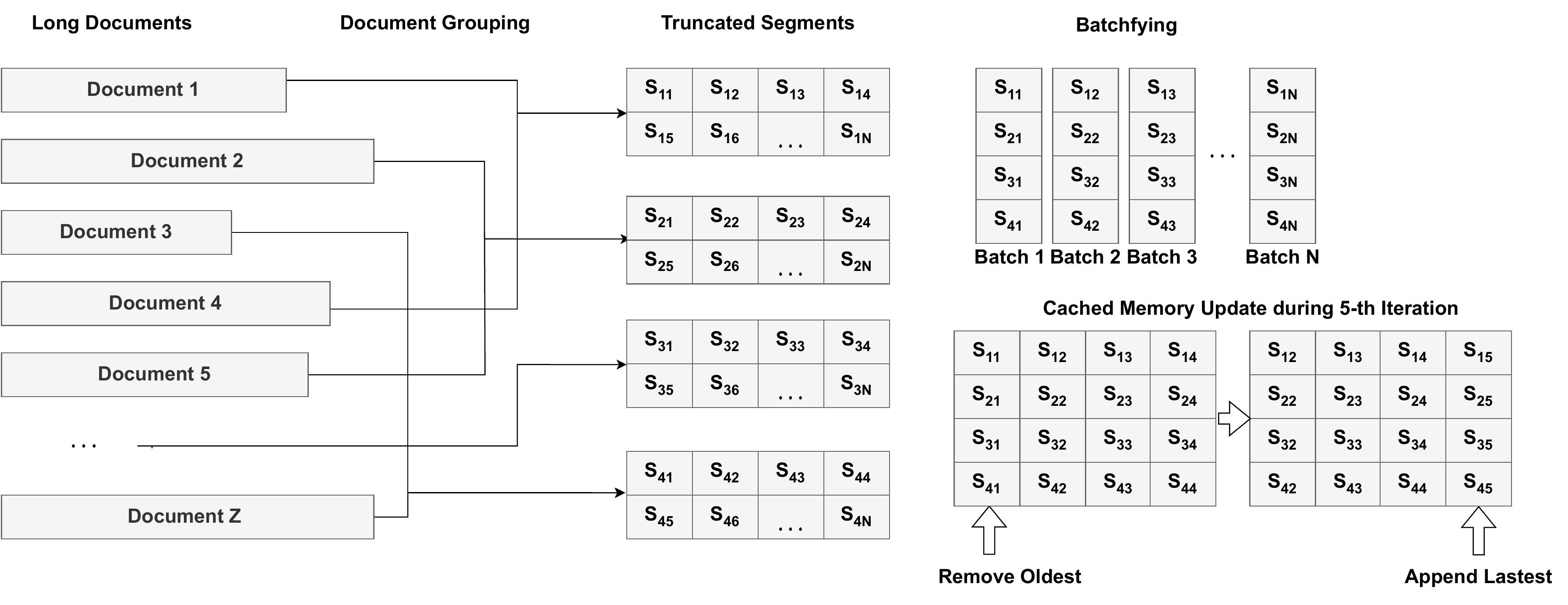} 
\caption{Batchfying the large text corpora into batches to ensure that each consecutive segments within each document is distributed in consecutive batches. 
}
\label{fig:batch}
\end{figure*}

\textbf{Batchfying the training corpora. } The conventional batchyfing process for large corpora truncates the whole corpora into consecutive fix-length text segments without padding and shuffles all segments to construct mini-batches~\citep{gpt2}. In contrast, \our{} must disable global shuffling and ensure the global causality at segment level. Firstly, we divide all long documents in training corpora into batch-size number of document groups with equivalent length and then perform a document-level shuffling within each group. Then, we concatenate shuffled documents within one group and truncate them into ordered segments. In order to ensure that two consecutive segments of one long document are distributed in two consecutive input batches after batchfying, we select one segment from batch-size number of document groups with the same inner-group index. Thus a mini-batch with batch-size number of segments are constructed from exactly batch-size number of document groups. In this way, as the training iteration steps, the cached attention key-value pairs in memory bank are exactly previous context of current inputs within the same document. The batchfying process is illustrated in Figure~\ref{fig:batch}.

\textbf{Training Corpus and Hyperparameters.} We sample a subset of the Pile~\citep{pile} as the training corpus, including BookCorpus2, Books3, OpenWebText2, Stack Exchange, Wikipedia, Gutenberg (PG-19), NIH ExPorter, and Pile-CC datasets. We reproduce GPT-2 (407M-params) as the pre-trained backbone LLM with Alibi~\citep{alibi} position embedding because original GPT-2~\citep{gpt2} adopts absolute position embedding, which is found to perform poorly to enable LLM to learn long-distance dependencies~\citep{transformerxl}. The backbone LLM holds a $L^\prime=24, H=16,d=64$ architecture. The SideNet holds a $L=12, H=16,d=64$ architecture. The training for memory-augmented adaptation iterates on 26B tokens, with a global 256 batch-size and 1024 sequence length. The chunk-size $csz$ is 4 tokens and the memory size $M$ is 65k key-value pairs of tokens. For each token, we retrieve $K$=64 attention key-value pairs for augmentation, which are $K/csz$=16 text chunks. The memory-augmentation layer is the 9-th layer of SideNet. The attention keys and values from 18-th layer of backbone LLM is cached into memory and used for future retrieval. Other training details are presented in Appendix C.

\textbf{Memory Retrieval Module.} The fixed memory-size of cached memory bank in one GPU is 65536 key-value pairs of tokens. We enable each GPU to construct and update their own memory retrieval module for efficiency. For the implementation of the efficient token-to-chunk retrieval, we use the \texttt{faiss}~\citep{Johnson2021BillionScaleSS} toolkit to construct an exact-search index on GPU to store the mean-pooled attention keys of text chunks and perform efficient retrieval. The \texttt{faiss} index maintains a fixed $M/csz$ keys and provides the efficient exact search w.r.t. inner product. The retrieval takes about 15ms per 1k tokens, which is 55\% timecost of backbone LLM forwarding pass. We can easily adapt the exact search index to approximate search index to gain more the retrieval efficiency. 

\textbf{Baselines.} In addition to the baseline of pre-trained GPT-2*, we reproduce Memorizing Transformer (MemTRM)~\citep{Wu2022MemorizingT} as another memory-augmented adaptation baseline. The MemTRM can be easily adapted to tune a pre-trained LLM to use external memory. We insert the knn-augmented layer proposed by MemTRM as the same 18-th layer in the LLM decoder. The MemTRM baseline is also trained for the same number of tokens under the same hyperparameter setting.

\begin{table*}[t]
\centering
\small
\scalebox{1}{
\begin{tabular}{@{}l | ccccc | c}
\hline    
\toprule
\textbf{Dataset} & \multicolumn{5}{c|}{\textbf{PG-22} } & \multirow{2}{*}{\textbf{ArXiv}} \\
\textbf{Splits} &  S1 & S2 & S3 & S4 & S5 & \\
 \midrule
\textbf{Len. Range} & \textsc{5k-10k}  & \textsc{10k-100k}  & \textsc{100k-500k} & \textsc{500k-1M} & \textsc{\textgreater 1M} & \textsc{\textless 60k} \\
\textbf{\#Documents}  &  500 & 100 & 30 & 8 & 1 & 100 \\
\textbf{Avg. \#tokens}  &  \textsc{7.6k}  & \textsc{47.6k}  & \textsc{140k} & \textsc{640k} & \textsc{1.2M} & \textsc{15.4k} \\
\bottomrule
\hline
\end{tabular}
}
\caption{Dataset Statistics of five splits of PG-22 based on length range and ArXiv. 
}
\label{table:pg22}
\end{table*}

\subsection{Long-Context Language Modeling}
\label{ssec:pastinputs}
 The long-context language modeling can easily benefit from the augmented decoupled memory of past long-contexts, in which the knowledge stored in retrieved attention key-values can play a useful role in providing significant background and contextual information to help models perform better on long-context language modeling. For instance, when trying to model a long-text book accurately, acquiring knowledge from previous background and character relationships can be helpful to model the consequent stories. 

\textbf{Evaluation Setting.}
We first compare \our{} and baselines on 3 long-context modeling datasets, \textit{Project Gutenberg 2020-2022}, \textit{ArXiv}, and \textit{ChapterBreak}. The majority of included books or papers in these datasets have the length of at least 16k tokens. All listed datasets are evaluated in \textbf{zero-shot} manner without any task-specific tuning. The detailed evaluation settings on 3 datasets are as follows:

\begin{itemize}[leftmargin=*]
\item \textbf{Project Gutenberg 2020-2022 Language Modeling Dataset.} We crawled and cleaned the books published between 2020 and 2022 under Project Gutenberg Library\footnote{https://www.gutenberg.org/} to build up a completely new long-text modeling dataset, named \textbf{PG-22}. It is highly differentiated from our training subset PG-19 in domains and writing styles, because books in PG-19~\citep{pg19} are published before 1919. We provide different validation splits of PG-22 based on length range, and data statistics are presented in Table~\ref{table:pg22}.
\item \textbf{ArXiv Dataset.} \texttt{ArXiv} dataset involves papers in the areas of Math, Computer Science, and Physics. We select a validation split of \texttt{ArXiv} paper subset in the Pile corpus~\citep{pile}. \texttt{ArXiv} subset of the Pile is excluded from our training and is an out-of-distribution dataset. We report the token-level language modeling perplexity on the long-context language modeling benchmarks of PG-22 and ArXiv. 
\item \textbf{ChapterBreak Benchmark.} \texttt{ChapterBreak} is proposed in \cite{ctb} as a challenging suffix identification dataset that requires LLMs to distinguish the beginning of the ground-truth next chapter from a set of hard negative segments sampled from the same book, given the long context of previous chapters. \texttt{ChapterBreak} requires processing global long-context to comprehend and identify the correct suffix. \cite{ctb} demonstrated that even state-of-the-art x-formers for long-text processing fail to effectively leverage long-range context to perform well on \texttt{ChapterBreak}. We select the Archive of Our Own (AO3) subset of \texttt{ChapterBreak} which contains fan-fictions extracted from AO3. \texttt{ChapterBreak} provides 8 splits based on the prefix length from 0.5k to 8k tokens to fit the length limit of different models. The splits of 4k, 6k, and 8k prefix are selected for evaluation. For LLMs that cannot process over 4k tokens, we abandon the front prefix to fulfill the maximum input length of LLMs. For MemTRM and \our{} model, we firstly load the given 4k/6k/8k prefix contexts into the cached memory and then do the scoring. we use the perplexity as the scorer for each candidate suffix segment in zero-shot evaluation manner. Then the suffix segment with lower perplexity is selected as the label. The suffix identification accuracy is used as the evaluation metric.
\end{itemize}

\begin{table*}[t]
\centering
\small
\scalebox{0.93}{
\begin{tabular}{@{}l cc | ccccc | c}
\hline
\toprule
\multirow{2}{*}{\textbf{Model}} & \textbf{In-Context} & \textbf{In-Memory} & \multicolumn{5}{c|}{\textbf{PG-22} } & \multirow{2}{*}{\textbf{ArXiv}} \\
& \textbf{Len.} & \textbf{Len.} &  \textsc{5k-10k}  & \textsc{10k-100k}  & \textsc{100k-500k} & \textsc{500k-1M} & \textsc{\textgreater 1M} \\
\midrule
GPT-2* & 1k& N/A & 22.78   &  24.39  &    24.12   & 24.97  &   18.07  & 11.05 \\ 
MemTRM & 1k & 65K & 21.77 & 23.56 &  23.23 & 24.16 &  17.39 &  10.81 \\
\midrule
\our{} & 1k & 65k & \textbf{21.29} &  \textbf{23.01} & \textbf{22.55} & \textbf{23.35} & \textbf{16.71} & \textbf{10.05} \\
\bottomrule
\hline
\end{tabular}
}
\caption{Evaluation results on long-context language modeling datasets. We report token-level perplexity (PPL) (lower the better) on all datasets. 
}
\label{table:lm}
\end{table*}

\begin{table*}[t]
\centering
\small
\scalebox{0.96}{
\begin{tabular}{@{}l c ccc c cc}
\hline    
\toprule
\multirow{2}{*}{\textbf{Model}} & \multirow{2}{*}{\textbf{\#Params}} & \textbf{In-Context} & \textbf{In-Memory} & \multicolumn{3}{c}{\textbf{ChapterBreak$_{\textbf{ao3}}$}} \\
 & & \textbf{Len.} & \textbf{Len.}& \textbf{ctx-4k} & \textbf{ctx-6k} & \textbf{ctx-8k}  \\
\midrule
GPT-2-XL$^\dagger$~\citep{gpt2} & 1.5B & 1K & N/A & 24\% &  24\% &  24\% \\
GPT-3$^\dagger$~\citep{gpt3} & 175B & 2K & N/A & 28\% & 28\%  & 28\%  \\
LocalTRM$^\dagger$~\citep{routingtrm} & 516M & 8K & N/A  & 24\% & 24\% & 24\% \\
RoutTRM$^\dagger$~\citep{routingtrm} & 490M & 8K & N/A  & 25\% & 24\% & 24\% \\
Bigbird$^\dagger$~\citep{bigbird} & 128M & 4K & N/A  & 26\% & 26\% & 26\% \\
\midrule
GPT-2* & 407M & 1K & N/A & 18.4\% & 18.4\% & 18.4\% \\ 
MemTRM & 407M & 1K & $\infty$ & 28.3\% &  28.7\% &  28.7\% \\
\midrule
\our{} & 558M & 1K & $\infty$ & \textbf{37.7\%} &  \textbf{39.4\%} & \textbf{40.5\%} \\ 
\bottomrule
\hline
\end{tabular}
}
\caption{Zero-shot Suffix Identification Accuracy on AO3 subset of \texttt{ChapterBreak}. Baselines marked with $^\dagger$ are directly cited from \cite{ctb}. The MemTRM and \our{} loads the given 4k/6k/8k prefix contexts into cached memory, while the input length to local context is still 1k tokens. }
\label{table:ctb}
\end{table*}

\textbf{Results.} The main results on evaluated long-context datasets are summarized in Table~\ref{table:lm}. The proposed \our{} model significantly outperform all considered baselines on long-text language modeling datasets, with improvements of -1.38 to -1.62 perplexity on different length splits of \textit{PG-22}, and -1.0 ppl on \textsc{ArXiv} datasets. Surprisingly, the proposed method achieves the state-of-the-art performance of 40.5\% accuracy on \texttt{ChapterBreak}$_\textnormal{AO3}$ suffix identification benchmark and outperforms both the strong long-context transformers and latest LLM GPT-3 with 313x larger parameters. The substantial improvements on these datasets demonstrate that \our{} can comprehend past long-context in cached memory to well complete the language modeling towards future inputs.

\subsection{Memory-Augmented In-Context Learning}
\label{ssec:incontext}

LLMs have the emerging capability of in-context learning (ICL) via learning knowledge non-parametrically from few-shot demonstration examples in the local context. However, conventional in-context learning is heavily restricted by input context length, rendering it ineffective to absorb supervision from sufficient demonstration examples in the training set. With the proposed unlimited-length memory augmentation, our \our{} method can overcome the limitation of the number of demonstration examples in the local context and even attend on the whole training set by loading it into the cached memory. In this way, \our{} goes beyond the conventional few-shot in-context learning and realized memory-augmented in-context learning with thousands of auxiliary demonstration examples.

\textbf{Evaluation Setting.} Here, we evaluate the in-context learning capability of baselines and the proposed \our{} model on five Natural Language Understanding (NLU) datasets, SST-2~\citep{socher2013recursive}, MPQA~\citep{wiebe2005annotating}, MR~\citep{auer2007dbpedia}, Subj~\citep{subj} and SST-5~\citep{socher2013recursive}. We evaluate models on two few-shot settings, 4-shot and 20-shot. The 4-shot demonstrations are data-insufficient scenario, while the 20-shot demonstrations can almost fulfill the 1k input length and provide sufficient contextual self-supervisions. We transform the k-shot examples to semantically meaningful demonstration examples via fixed text template, i.e., $d_i$="Review: $x_i$ Sentiment: $y_i$",$\forall \{(x_i,y_i)\}_{i=1}^k\in \mathcal{D}_{\textnormal{train}} $ for sentiment analysis tasks. Additionally, we evaluate the 3-shot ICL on question-answering tasks of SQuAD~\citep{squad} under an open-ended generation setting. The details of all prompt templates are presented in Appendix D. Then we concatenate the demonstration examples with newlines to delimit them. The prediction label is directly generated using greedy decoding given the demonstration examples and test cases in context. The prediction accuracy is used as the evaluation metric. We report the mean and standard deviation of 6 runs with different random seeds to overcome the randomness in selecting k-shot demonstration examples. As illustrated before, the chunk size controls the granularity of retrieved text chunks. As the select NLU datasets require to retrieve fine-grained labels from cached memory, we perform an hypperparameter selection on the validation set of SST-2, and the best chunk-size 2 is used to report the results for MemTRM and our model.

\textbf{Results.} The results on in-context learning are summarized in Table~\ref{table:NLU} and Table~\ref{tab:squad}.
\our{} achieves remarkable improvements on all NLU tasks in 20-shot sufficient in-context setting, with +8.0 average scores increase over pretrained GPT-2* and MemTRM. Meanwhile, \our{} also brings performance improvements on the scenario of 4-shot demonstrations in local context. 
Additionally, \our{} improves the in-context learning capabilities of LLMs on open-ended generation tasks, with +4.5 EM score increase on SQuAD. 
\begin{wraptable}{r}{5.5cm}
\centering
\small
\resizebox{.4\textwidth}{!}{
\begin{tabular}{lcc}
\toprule
\textbf{Model} & \textbf{EM} & \textbf{F1}\\
\midrule
GPT-2* & 22.28$_{2.3}$ & 30.78$_{2.0}$ \\
MemTRM & 22.84$_{3.5}$ & 32.65$_{2.8}$  \\
\our{} & 26.77$_{2.3}$ &  35.70$_{2.0}$ \\
\bottomrule
\end{tabular}
}
\caption{Exact match (EM) and F1 scores of 3-shot (about 1k tokens) in-context learning on SQuAD. \our{} loads 200 extra demonstration examples into cached memory.
}
\label{tab:squad}
\end{wraptable}
The results indicate that the demonstration examples loaded in cached memory can be regarded as auxiliary contextual demonstrations to attend to and be helpful for in-context learning. 
\our{} model can harvest both the task-relevant knowledge in both local contextual demonstrations and in-memory augmented demonstrations for better in-context learning.

\begin{table*}[t]
\small
\centering
\scalebox{1}{
\begin{tabular}{@{}lcccccccc}
\hline    
\toprule
\multirow{2}{*}{\textbf{Model}} & {\textbf{In-Context}} & {\textbf{In-Memory}} & \textbf{SST-2} & \textbf{MR} & \textbf{Subj} & \textbf{SST-5} &\textbf{MPQA} & \multirow{2}{*}{\textbf{Avg.}}   \\
 & \textbf{\#Demons.} & \textbf{\#Demons.} & ACC$\uparrow$& ACC$\uparrow$& ACC$\uparrow$ & ACC$\uparrow$& ACC$\uparrow$ &  \\
\midrule
Majority & N/A & N/A & 50.9 & 50.0 & 50.0 & 20.0 & 50.0 & 44.2 \\
\midrule
GPT-2* & 4 & N/A & 68.3$_{11.6}$ & 64.7$_{12.5}$ & 51.9$_{4.2}$ & 31.4$_{4.4}$ & 61.5$_{11.8}$ & 55.6 \\
MemTRM & 4 & 2000 & 67.5$_{12.4}$ & 64.6$_{11.3}$ & 53.2$_{6.0}$  & 29.6$_{4.4}$ & 63.0$_{12.1}$ & 55.6  \\
\our{} & 4 & 2000 & \textbf{71.8}$_{14.0}$ & \textbf{65.1}$_{11.0}$ & \textbf{53.8}$_{3.7}$ & \textbf{36.0}$_{6.8}$ & \textbf{65.4}$_{12.8}$ & \textbf{58.4}\\
\midrule
GPT-2* & 20 & N/A &  68.2$_{11.5}$ & 63.4$_{5.2}$ & 57.6$_{10.2}$ & 33.6$_{6.0}$ & 70.8$_{7.6}$ & 58.7 \\
MemTRM & 20 & 2000 & 65.1$_{9.6}$ & 65.1$_{9.3}$ & 58.2$_{10.6}$ & 31.9$_{6.3}$ 
& 72.7$_{7.4}$ & 58.6 \\
\our{} & 20 & 2000 & \textbf{78.0}$_{14.1}$ & \textbf{78.6}$_{3.3}$ & \textbf{65.6}$_{8.5}$ & \textbf{36.5}$_{7.5}$ & \textbf{74.6}$_{7.3}$ & \textbf{66.7} \\
\bottomrule
\hline
\end{tabular}
}
\caption{Accuracy [\%] of 4-shot and 20-shot ICL on 5 NLU tasks (SST-2, mr, subj, SST-5, mpqa). We sample 2000 extra demonstration examples and load them into cached memory. The subscript is the standard deviation across 6 runs. Avg. refers to the average accuracy on 5 datasets.
}
\label{table:NLU}
\end{table*}

\subsection{Ablation Studies}

So far, we empirically verify the effectiveness and superiority of \our{} in utilizing cached memory for long-context modeling, long-context understanding, and many-shot in-context learning. As the design of cached memory bank involves many hyperparameters like memory size $msz$ and chunk-size $csz$, we perform a series of ablation studies to evaluate the effects of these hyperparameters on task performance. 

\begin{figure}[t]
\centering
\subfigure[]
{
	\begin{minipage}{0.5\textwidth}
	\centering          
	\includegraphics[width=1\textwidth]{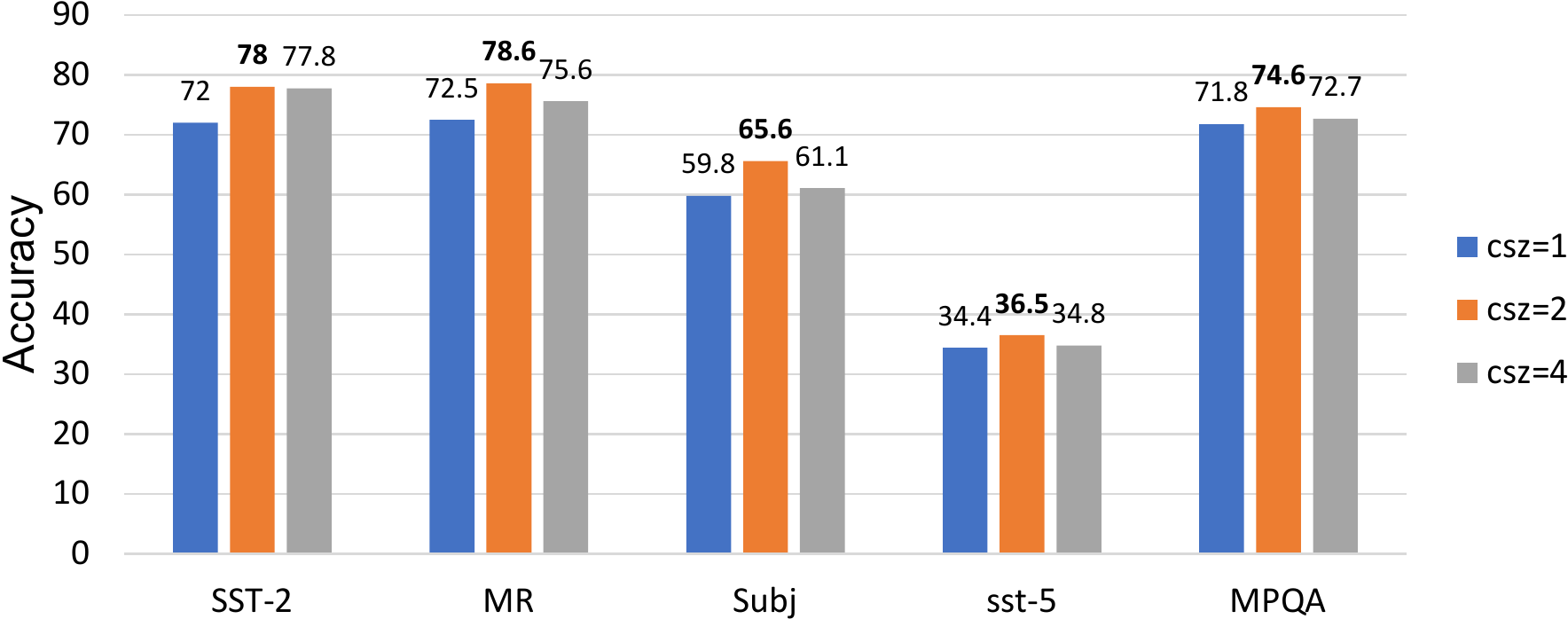}
    \label{fig:csz}
	\end{minipage}
}
\subfigure[]
{
	\begin{minipage}{0.45\textwidth}
	\centering     
	\includegraphics[width=1\textwidth]{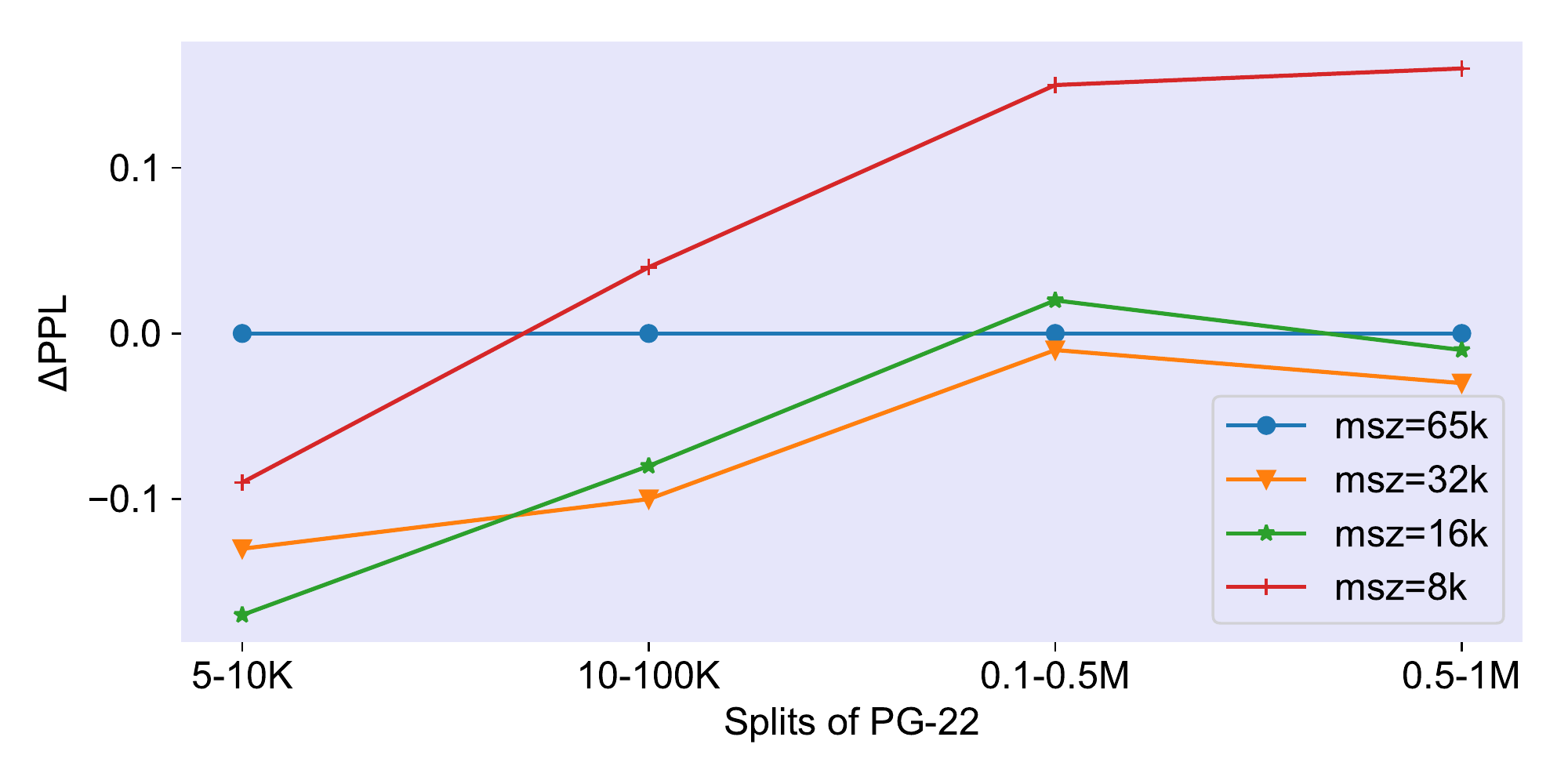}
    \label{fig:msz}
	\end{minipage}
}
\vspace{-10pt}
\caption{(a) Accuracy on 5 NLU datasets given different chunk size during inference; (b) $\Delta$Perplexity on 4 splits of PG-22 given different memory size during inference, in which the perplexity when $msz$=65k is used as baseline.}
\label{fig:ablation} 
\end{figure}

\textbf{Effects of Chunk-Size.} As analyzed before, the chunk-size $csz$ controls the granularity of retrieval and thus it may make a difference to tasks with requirements of fine-grained retrieval like in-context learning. We perform an ablation study on the effects of various chunk-size $csz\in \{2,4,8\}$ on in-context learning and the results are presented in Figure~\ref{fig:csz}. The chunk size of 2 yields the best performance on in-context learning tasks on five NLU datasets, which is consistent with the property of NLU tasks with the requirement of fine-grained retrieval and fusion towards classification label tokens.

\textbf{Effects of Memory Size.} The memory size (msz) controls the capacity of the memory bank. In general, the memory size should be compatible with the average length of documents or contexts, \ie, a set of books with average 16k tokens should deploy the memory size of 16k tokens in cached memory. The training $msz$ of 65 tokens is excessive for downstream tasks such as ChapterBreak as the whole prefix context length does not exceed 65k tokens. Thus, we perform an ablation study on the effects of memory size $msz\in\{8k,16k,32k,65k\}$ during the inference stage on the PG-22 language modeling datasets and the results are shown in Figure~\ref{fig:msz}. To model the books with average 8k-50k length, the smaller memory size $16k$ which is consistent with the average length of target books yields the best perplexity.

%% file: 4_related_work.tex
\vspace{-5pt}
\section{Related Work}
\label{sec:related}
\textbf{Large Language Models.} Large Language Models, \ie GPT-2~\citep{gpt2}, GPT-3~\citep{gpt3}, OPT~\citep{zhang2022opt}, and BLOOM~\citep{scao2022bloom}, significantly revolutionized NLP research and promoted the state-of-the-art of various language understanding, language generation~\citep{wang2022task}, and even vision-language tasks~\citep{valm}. Additionally, via scaling the model parameters, LLMs exhibit ``emergent abilities``~\citep{wei2022emergent} like few-shot in-context learning~\citep{gpt3}, multi-step reasoning~\citep{cot}, code completion, etc. 

\textbf{x-formers.} To enable transformers to attend on longer context, many variants of ``x-formers`` are proposed. Transformer-XL~\citep{transformerxl} proposes to cache attention keys and values of past segment and reuse them in recurrent manner. Recent seminal works of x-formers, including LinFormer~\citep{wang2020linformer}, LongFormer~\citep{beltagy2020longformer}, Routing Transformer~\citep{routingtrm}, proposed various sparse attention mechanisms for decreasing $O(n^2)$ complexity to $O(n\log n)$ or even $O(n)$. BigBird~\citep{bigbird} achieves a 4k sequence length via attending on a subset of context tokens. Although these x-formers achieve substantial efficiency improvements, such efficiency gains are not remarkable when modeling sequences that spans book-level length. Moreover, the largest sequence length of these methods is still upper-bounded by 16k tokens, making them invalid in modeling long-sequences at the book or wikipedia-page level (\ie average 70k tokens for full-length books in PG19 dataset~\citep{pg19}). 

\textbf{Side-Tuning.} The method of Side-Tuning~\citep{sidetuning,lst} is a task-specific tuning method for pre-trained models via training a lightweight side-network that is fused with the fixed pre-trained network via summation. Our method inherits the idea of adopting a side-network but distinguishes the side-tuning method in terms of learning objective and cross-network fusion ways. \our{} proposes to augment LLMs with decoupled memory for memorizing long past inputs, which does not involve any task-specific tuning. The cross-network residual connections proposed by \our{} is novel and distincts from the vanilla summation of Side-Tuning.

%% file: 5_conclusion.tex
\vspace{-5pt}
\section{Conclusion}
\label{sec:conclusion}

In this paper, we propose to augment LLMs with long-term memory for enabling them to memorize long-form context and gain long-form memory. The designed decoupled memory module can cache attention key and value pairs of past inputs for future retrieval and fusion. A decoupled residual SideNet is introduced as the memory retriever and reader, meanwhile the LLM itself is frozen and works as knowledge and memory encoder. Experiments on various long-contextual language modeling datasets demonstrate the effectiveness of our model over other memory-augmentation baselines. The proposed method can also enable in-context learning of LLMs to overcome the limited number of demonstration examples in context, which is constrained by the contextual length, via caching thousands of auxiliary demonstration examples in memory. 


%% file: supplementary.tex
\section{Inference Efficiency and GPU-Memory Efficiency}
When the model is required to comprehend long sequences, the proposed method \our{} can load the out-of-boundary inputs into the cached memory as previous context. Thus, the memory usage and inference speed can be significantly improved compared with vanilla self-attention-based models. The detailed statistics in terms of the efficiency is presented in Table~\ref{table:efficiency}.

\begin{table*}[h]
\centering
\small
\scalebox{1}{
\begin{tabular}{@{}l cc | ccc}
\hline
\toprule
\multirow{2}{*}{\textbf{Model}} & \textbf{In-Context} & \textbf{In-Memory} & \textbf{Inference Speed}  & \textbf{GPU-Memory Usage}  \\
& \textbf{Len.} & \textbf{Len.} & (tokens/s)$\uparrow$ & (MBs)$\downarrow$ \\
\midrule
GPT-2* & 4k & N/A & 14666 & 20671 \\ 
\our{} & 1k & 3k & 22638 & 13335\\
\midrule
GPT-2* & 8k & N/A & 8417 & 54195	\\ 
\our{} & 1k & 7k & 21343 & 13437\\
\bottomrule
\hline
\end{tabular}
}
\caption{The superiority of our method over fully dense self-attention (GPT-2*) in terms of inference speed and GPU-memory utilization.
}
\vspace{-10pt}
\label{table:efficiency}
\end{table*}

\section{Training Details}
\label{sup:train}
The pre-training of reproduced GPT-2* iterates on 117B tokens in total, with 512 batch-size and 1024-token fixed segment-length. The Adam optimizer~\citep{Kingma2015AdamAM} is adopted in memory-augmented adaptation training. The pre-training and adaptation are trained on 16 32GB-Tesla-V100 GPUs. Other detailed training hypperparamters and settings are presented in Table~\ref{table:hp}.

\begin{table}[ht]
\begin{center}
\small
    \begin{tabular}{l|c}
    \hline
    \toprule
    \textbf{Hyperparameter}  & \our{} \\ 
    \midrule
    \multicolumn{2}{c}{\textbf{Reproduced GPT-2* Backbone LLM Hyperparameters}} \\
    Parameters & 407M \\  
    Precision & \texttt{float16}  \\ 
    Layers   & 24 \\ 
    Hidden dim. & 1024  \\ 
    Attention heads & 16  \\
    Head Dim & 64 \\
    Vocab size & 52k \\ 
    Sequence length & 1024 \\
    Position emb. & \texttt{Alibi} \\ 
    Tied embedding & \texttt{False} \\ 
    \midrule
    \multicolumn{2}{c}{\textbf{SideNet Hyperparameters}} \\
    Parameters & 151M \\  
    Precision & \texttt{float16}  \\ 
    Layers   & 12 \\ 
    Hidden dim. & 1024  \\ 
    Attention heads & 16  \\
    Head Dim & 64 \\
    Sequence length & 1024 \\
    \midrule
    \multicolumn{2}{c}{\textbf{Memory-Augmented Adaptation Hyperparameters}} \\
    Global Batch Size  & 256 \\ 
    Learning rate & 2.0e-4  \\ 
    Total tokens & 26B \\ 
    Warmup tokens & 0 \\ 
    LR Decay style & polynomial \\ 
    Adam $(\beta_1, \beta_2)$ & (0.9, 0.98) \\   
    Adam eps & 1e-06 \\
    Weight decay & 0.01 \\ 
    Gradient clipping & 2.0 \\
    \bottomrule
    \hline
    \end{tabular}
    \vspace{5pt}
\caption{Memory-Augmented Adaptation and Architectural Hyperparameters.}
\label{table:hp}
\end{center}
\end{table}

\section{Prompting Templates}
\label{sec:template}
We present all hand-crafted in-context learning prompting templates and labels for 5 NLU datasets and Squad QA dataset in Tabel~\ref{table:prompt}.

\begin{table*}[h]
    \centering
    \small
    \scalebox{1}{
        \begin{tabular}{ l |l l}
            \hline    
            \toprule
            \textbf{Task} & \textbf{Prompt} & \textbf{Labels} \\
            \midrule
            \textbf{SST-2} & Review: [Sentence] Sentiment: [Label] & \{positive, negative\} \\
            \midrule
            \textbf{MR} & Review: [Sentence] Sentiment: [Label] & \{positive, negative\} \\
            \midrule
            \textbf{MPQA} & Review: [Sentence] Sentiment: [Label] & \{positive, negative\}\\
            \midrule
            \textbf{SST-5} & input: [Sentence] type: [Label] & \{terrible,bad,okay,good,great\} \\
            \midrule
            \textbf{Subj} & input: [Sentence] type: [Label] &  \{objective, subjective\} \\
            \midrule
            \textbf{Squad} & Passage: [Passage]$\backslash$n Question: [Question] Answer: [Answer] \\
            \bottomrule
            \hline
            \end{tabular}
        }
    \caption{The hand-crafted prompts used to query the model predictions on the zero-shot evaluation of 5 NLU datasets and one question-answering dataset Squad.
    }
    \label{table:prompt}
\end{table*}